%% file: main.tex
\documentclass[journal]{IEEEtran}
\usepackage{orcidlink}

\usepackage[ruled,norelsize,linesnumbered]{algorithm2e}

\ifCLASSOPTIONcompsoc
  \usepackage[caption=false,font=normalsize,labelfont=sf,textfont=sf]{subfig}
\else
  \usepackage[caption=false,font=footnotesize]{subfig}
\fi
\usepackage{url}

\makeatletter
\newcommand{\removelatexerror}{\let\@latex@error\@gobble}
\makeatother

\newcommand{\blue}[1]{{\color{black} #1}}

\begin{document}

\title{Mobiprox: Supporting Dynamic Approximate Computing on Mobiles}

\author{Matevž Fabjančič\orcidlink{0009-0006-9269-141X}\thanks{M. Fabjančič is with Cosylab, Ljubljana, Slovenia},
Octavian Machidon\orcidlink{0000-0003-3133-1008},\thanks{O. Machidon is with Faculty of Computer and Information Science, University of Ljubljana, Slovenia}
Hashim Sharif\orcidlink{0000-0002-9496-9028}\thanks{H. Sharif is with Department of Computer Science at the University of Illinois at Urbana-Champaign, USA},
Yifan Zhao\orcidlink{0009-0007-7080-891X},\thanks{Y. Zhao is with Department of Computer Science at the University of Illinois at Urbana-Champaign, USA}
Saša Misailović\orcidlink{0000-0001-7319-8845}\thanks{S. Misailović is with Department of Computer Science at the University of Illinois at Urbana-Champaign, USA}, and~Veljko Pejović\orcidlink{0000-0002-9009-0024}
\thanks{V. Pejović is with Faculty of Computer and Information Science, University of Ljubljana, Slovenia and Computer Systems Department, Institute Jozef Stefan, Ljubljana, Slovenia (email:veljko.pejovic@fri.uni-lj.si)}
}


\maketitle

\begin{abstract}

Runtime-tunable context-dependent network compression would make mobile deep learning (DL) adaptable to often varying resource availability, input ``difficulty'', or user needs. The existing compression techniques significantly reduce the memory, processing, and energy tax of DL, yet, the resulting models tend to be permanently impaired, sacrificing the inference power for reduced resource usage. The existing tunable compression approaches, on the other hand, require expensive re-training, do not support arbitrary strategies for adapting the compression and do not provide mobile-ready implementations. 

In this paper we present Mobiprox, a framework enabling mobile DL with flexible precision. Mobiprox implements tunable approximations of tensor operations and enables runtime-adaptable approximation of individual network layers. A profiler and a tuner included with Mobiprox identify the most promising neural network approximation configurations leading to the desired inference quality with the minimal use of resources. Furthermore, we develop control strategies that depending on contextual factors, such as the input data difficulty, dynamically adjust the approximation levels across a mobile DL model's layers. We implement Mobiprox in Android OS and through experiments in diverse mobile domains, including human activity recognition and spoken keyword detection, demonstrate that it can save up to 15\% system-wide energy with a minimal impact on the inference accuracy.

\end{abstract}




\newcommand{\mnCifar}{\texttt{mobilenet\_cifar10}}
\newcommand{\mnUci}{\texttt{mobilenet\_uci-har}}
\newcommand{\anCifar}{\texttt{alexnet2\_cifar10}}
\newcommand{\vggCifar}{\texttt{vgg16\_cifar10}}
\newcommand{\resUci}{\texttt{resnet50\_uci-har}}

\IEEEpeerreviewmaketitle

\input{tex/introduction}

\input{tex/related_work}

\input{tex/preliminaries}

\input{tex/mobiprox}

\input{tex/strategies}

\input{tex/methodology}

\input{tex/evaluation}

\input{tex/discussion}
\input{tex/conclusion}

\section{Acknowledgements}
This work was partly supported by the Slovenian Research Agency (grants no. P2-0098, P2-0426, N2-0136 ``Bringing Resource Efficiency to Smartphones with Approximate Computing'' and J2-3047 ``Context-aware on-device approximate computing'').

\bibliographystyle{IEEEtran}
\bibliography{sample-base}

\begin{IEEEbiography}[{\includegraphics[width=1in,height=1.25in,clip,keepaspectratio]{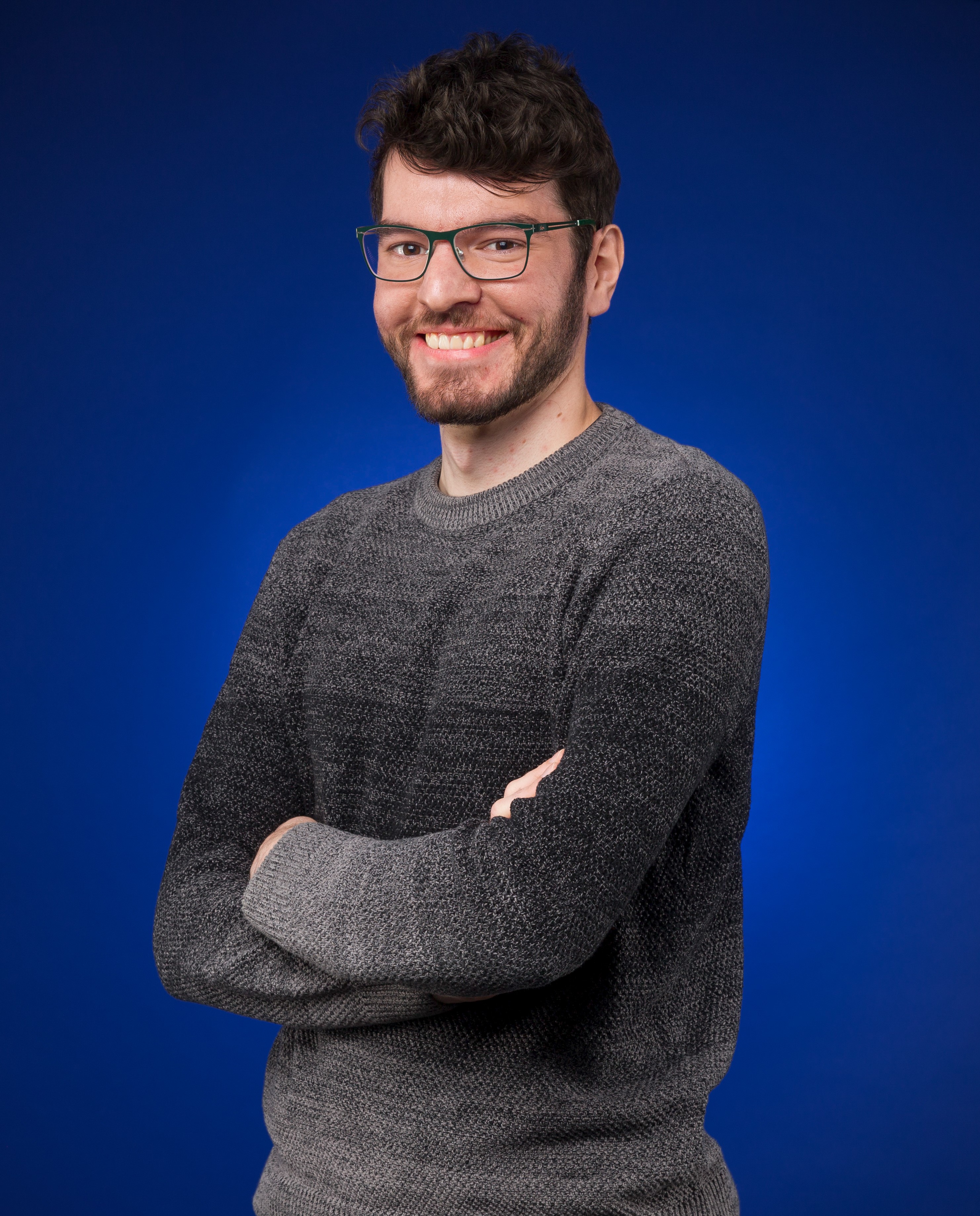}}]{Matevž Fabjančič} is a member of Cosylab's Medical Services department, where he develops software that supports the process of cancer treatment through radiotherapy. He graduated with a master's degree from the University of Ljubljana, Faculty of Computer and Information Science in 2021. In 2022, he was awarded the Prešeren Award for Students of University of Ljubljana for his master's thesis. Contact him at  matevz.fabjancic@gmail.com.

\end{IEEEbiography}

\begin{IEEEbiography}[{\includegraphics[width=1in,height=1.25in,clip,keepaspectratio]{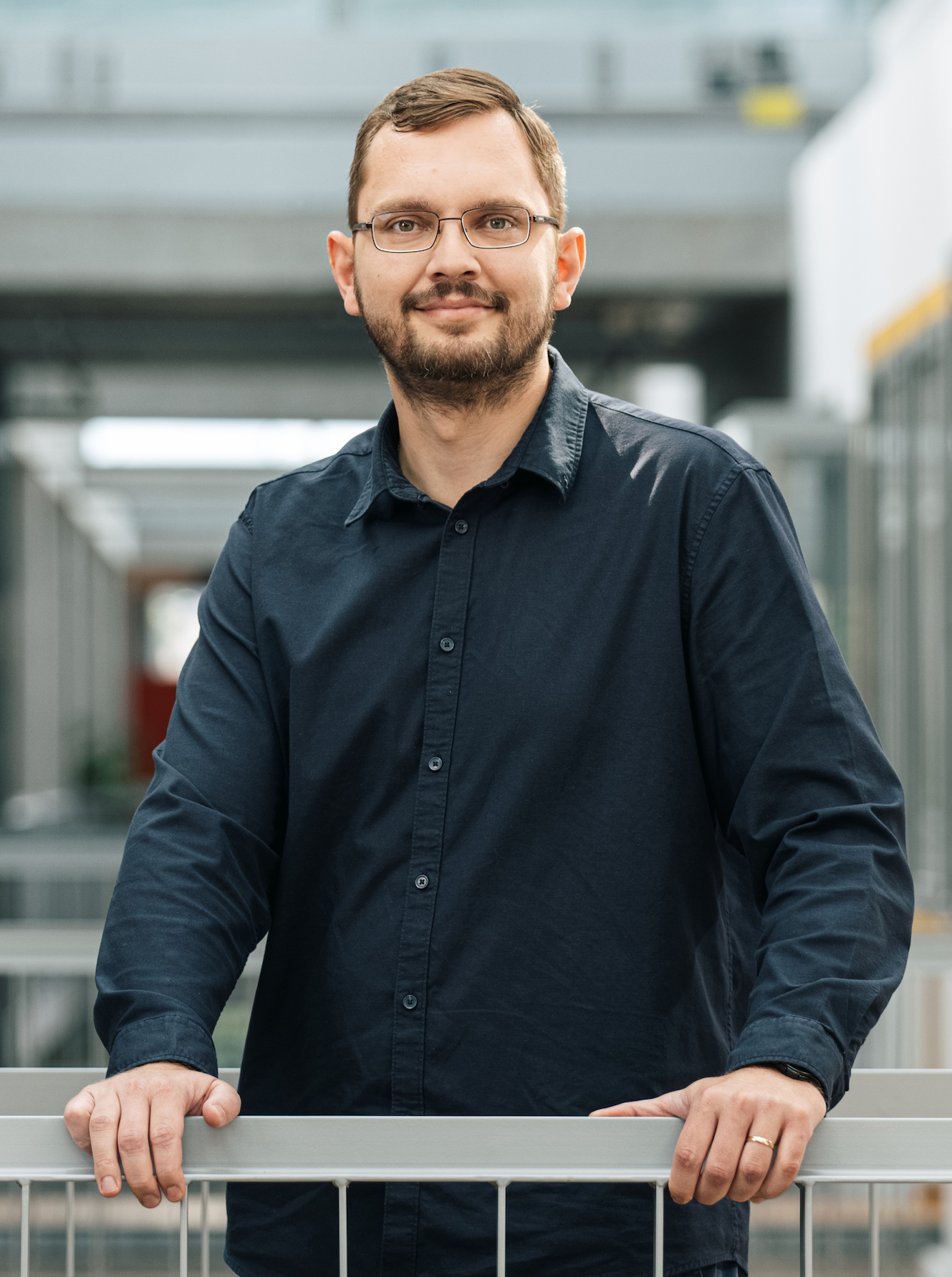}}]{Octavian Machidon} is an assistant professor at the Faculty of Computer and Information Science, University of Ljubljana, Slovenia. He received his PhD in reconfigurable computing from Transilvania University of Brasov, Romania. In Ljubljana, he is currently focused on implementing approximate mobile computing solutions for enabling energy-efficient mobile applications. Contact him at octavian.machidon@fri.uni-lj.si.
\end{IEEEbiography}

\begin{IEEEbiography}[{\includegraphics[width=1in,height=1.25in,clip,keepaspectratio]{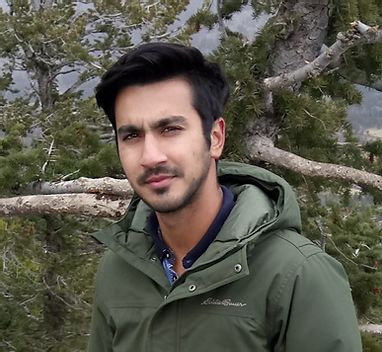}}] {Hashim Sharif} is a research scientist at AMD Research. He completed his PhD and Postdoc in Computer Science at the University of Illinois at Urbana-Champaign. He works in the areas of Compilers and Systems for Machine Learning. Contact him at hsharif3@illinois.edu.

\end{IEEEbiography}

\begin{IEEEbiography}[{\includegraphics[width=1in,height=1.25in,clip,keepaspectratio]{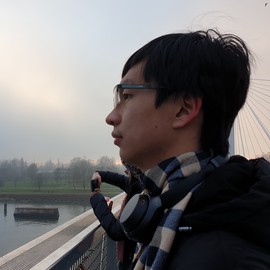}}]{Yifan Zhao} is a Ph.D. student at the University of Illinois at Urbana-Champaign, USA. He works in the area of Compilers and Systems for Machine Learning, specifically accuracy-aware optimizations and autotuning techniques for better discovering approximation choices. Contact him at yifanz16@illinois.edu.

\end{IEEEbiography}

\begin{IEEEbiography}[{\includegraphics[width=1in,height=1.25in,clip,keepaspectratio]{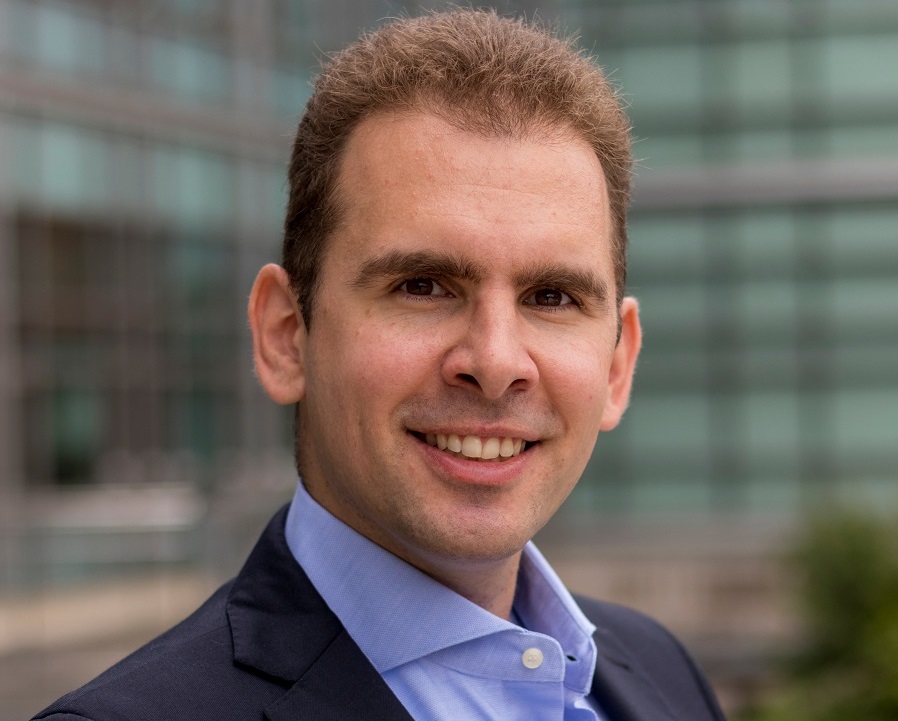}}]{Saša Misailović} is an Associate Professor in the Department of Computer Science at the University of Illinois at Urbana-Champaign. He received his PhD from MIT in Summer 2015. His research interests include programming languages, compilers, and software engineering, with an emphasis on improving performance, energy efficiency, and resilience in the face of software errors and approximation opportunities. He was recognized with NSF CAREER Award and multiple best paper awards. Contact him at misailo@illinois.edu
\end{IEEEbiography}

\begin{IEEEbiography}[{\includegraphics[width=1in,height=1.25in,clip,keepaspectratio]{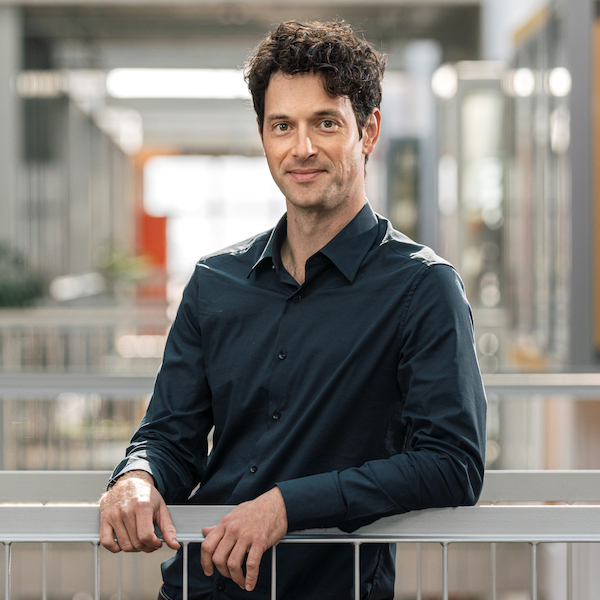}}]{Veljko Pejović} is an associate professor at the Faculty of Computer and Information Science, University of Ljubljana, Slovenia. He received his PhD in computer science from the University of California Santa Barbara and was a Research Fellow at the Computer Science Department, University of Birmingham, UK. 
His research focuses on resource-efficient mobile systems, human-computer interaction, and cybersecurity in ubiquitous systems. His awards include the best paper nomination at ACM UbiComp, best paper runner-up at IEEE Pervasive Computing, and the first prize at Orange D4D challenge for his work on epidemics modeling. 
Contact him at veljko.pejovic@fri.uni-lj.si
\end{IEEEbiography}

\end{document}

%% file: tex/introduction.tex
\section{Introduction}
\label{sec:introduction}

\IEEEPARstart{P}{owerful} services enabled by deep learning, such as real-time camera-based object detection, online translation, and human activity recognition (HAR), are becoming increasingly available on mobile devices. Indeed, DL is an integral part of more than 12\% of application downloads from the Android store platform~\cite{10.1145/3308558.3313591}. However, the new affordances do not come for free -- large DL models may overload the limited memory of mobile devices, the computational burden may lead to significant delays before the results are available, and the power needed for processing may quickly deplete the mobile's battery. 

Reducing the complexity of neural networks (NNs) is the primary means of making DL mobile friendly. Such complexity reduction may be inherent to the network design -- MobileNet~\cite{howard2017mobilenets}, EfficientNet~\cite{tan2019efficientnet}, and ShuffleNet~\cite{zhang2018shufflenet} represent some of the architectures that are specifically designed for the mobile's limited memory resources. Yet, the computational burden of these networks may still be overwhelming for a wide range of heterogeneous edge devices~\cite{almeida2019embench}.
Both memory and computational complexity can be further reduced by a gamut of NN compression techniques. These include parameter quantization~\cite{wu2016quantized}, weight pruning~\cite{niu2020patdnn}, NN distillation~\cite{hinton2015distilling},  to name a few. \blue{The key issue with such complexity reduction is that the network parameters are permanently changed. Thus, in case that the resulting inference accuracy is reduced, that reduction remains permanent.}

On mobiles, on the other hand, \textit{deep learning compression needs to be adaptable to the context of use}: a compressed model that reliably recognises a user's speech commands when used in a quiet indoor location, might completely fail in noisy outdoor environments; similarly, a user might tolerate a more compressed model that occasionally misclassifies her physical activity during her daily routine, but would require a more accurate model while exercising. A rigid approach to DL compression is against the often dynamic nature of mobile computing, where both a user's requirements with respect to the result accuracy~\cite{machidon2020watching}\blue{, as} well as the difficulty of given NN input~\cite{machidon2021queen}, may vary as the context of use changes. 

Recently, proposals have been made to enable dynamic accuracy/complexity adaptation of NNs. Examples include dynamic quantization by AnyPrecision~\cite{yu2020any}, dynamic adjustment of layer width through Slimmable Neural Networks~\cite{yu2018slimmable}, or dynamic pruning proposed in~\cite{10.5555/3294771.3294979}. Common for all of the above \blue{dynamic adaptation} approaches is that they do not support prebuilt networks, but require specialized training that can take days or weeks for large datasets and architectures before real-time adaptation can be used. Furthermore, despite targeting dynamic environments, the above works do not actually provide mobile-ready implementations. Translating the benefits provided by high-level demonstrations (often implemented in PyTorch) to mobile energy savings requires significant engineering effort, as modern mobile DL frameworks such as TensorFlow Lite do not support the versatility of high-level frameworks such as PyTorch.

Advances in a different research area -- compilers for heterogeneous systems -- have recently addressed the issue of ''optimal'' NN compilation, where individual tensor operations are implemented in accordance with underlying hardware capabilities. Along these lines,  ApproxHPVM~\cite{sharif2019approxhpvm} enables the execution of convolutional neural network (CNN) operations with varying degrees of approximation, provided that the hardware/OS supports approximate computation.
However, ApproxHPVM targets server environments, generates only CUDA-ready binaries, and does not support compilation for mobile hardware (Android or iOS). With the help of ApproxTuner~\cite{sharif2021approxtuner}, approximation levels within ApproxHPVM can be dynamically adapted, yet, the provided adaptation method is simple, reactive and context-oblivious.


In this paper we present \textbf{Mobiprox} -- a novel framework that enables context-adaptable approximation of DL operations executed on the mobile platform.
Our guiding vision is that \textit{data scientists are not mobile system experts}. Therefore, deep learning modeling should be disentangled from system-level performance optimization. Mobiprox aims to support efficient on-device execution of an arbitrary pre-trained mobile network architecture. Furthermore, we do not require that a developer knows which optimizations (in our case -- execution approximations) are available on the device. Still, we give a developer an option of (dynamically) setting an operational point along the inference accuracy vs. resource usage trade-off curve, yet, in the limit case, the developer need not even set this point, but merely let Mobiprox tune the execution according to its internal approximation adaptation algorithms.

We implement Mobiprox at low levels of the computing stack to support a wide range of NN architectures and embrace various approximation techniques exposed by the underlying hardware and the OS\footnote{The specific implementation presented in this paper supports perforated convolution, filter sampling, and half-precision quantisation.}. To support context-sensitive runtime adaptation Mobiprox identifies Pareto-optimal approximation configurations during the off-line tuning stage. The system then enables the network to glide across different speedup/accuracy trade-off points during the runtime. The key novelty of Mobiprox are also the adaptation algorithms that guide the runtime approximation adaptation according to a given goal, e.g. maximal energy savings. 


With Mobiprox, we address multiple challenges that stand in the way towards adaptable \mbox{approximate mobile DL:}
\begin{itemize} 

\item \textbf{The difficulty of implementing approximate operations at an appropriate level of the mobile computing stack;} Over the past two decades, a number of approximate computing techniques have been developed -- from approximate adders and multipliers to loop perforation and task skipping~\cite{mittal2016survey}. Because of their small form factor, however, mobile devices can rarely accommodate both approximate and accurate versions of hardware circuits. Software techniques, on the other hand, often require strong developer involvement, e.g., in marking loops eligible for perforation. Therefore, we focus on software-level approximation of tensor operations. Since mobile DL frameworks (e.g., TensorFlow Lite) and even libraries of specialized functions for mobile DL (e.g., ARM Compute Library) aggregate tensor operations, we implement both approximate and precise tensor operations from basic linear algebra subprogram (BLAS) primitives;

\item \textbf{The issue of modifying neural network operation at runtime on a mobile device;} mobile DL frameworks do not support dynamic graph reconfiguration, thus even the existing dynamic approximation schemes (such as Slimmable Neural Networks~\cite{yu2018slimmable}) do not work on mobiles; to overcome this limitation, we implemented our custom approximations at a fine-grained level and exposed the calls for setting the approximation level at runtime through Java Native Interface;

\item \textbf{The lack of algorithms and tools for context-aware adaptation of mobile DL;} a certain classification accuracy level might be acceptable in some situations, but not in others; in addition, an approximated DL model that works well for certain inputs, might not provide correct classification for some other inputs; finally, gauging model performance at runtime is challenging; we first devise proxies for measuring classification performance (the same-class instance counting-based and the softmax confidence-based) and then develop algorithms (state-based, and confidence-based) for dynamically adapting the approximation.


\end{itemize}

\noindent Towards this end, we present the following contributions:
\begin{itemize}
    \item \textbf{We develop an end-to-end approximate configuration search, selection, and compilation pipeline for mobile devices.} Our solution integrates state-of-the-art heterogeneous compilation infrastructure, approximate configuration search framework, and a widely used LLVM compiler into an Android-ready pipeline; furthermore, our solution supports dynamic configuration loading;
    \item \textbf{We devise novel strategies for runtime approximation configuration adaptation}; based on the problem properties or the classifier confidence, our solutions 
    ensure that the desired inference accuracy is achieved with the minimal use of a mobile's resources;
    \item \textbf{We implement selected approximate computing primitives at a low-level of the mobile computing stack}, supporting both on-CPU and on-GPU approximate execution of different tensor operations for mobile devices.
    \item \textbf{Our evaluation shows that Mobiprox brings substantial energy savings while preserving the classification accuracy.}  We perform experiments on both a \blue{single-board computer} (for precise energy measurements) and \blue{on commodity smartphones performing real-time inference}, using different NN architectures and multiple application domains, including human activity classification and spoken keyword recognition. Our evaluation demonstrates that, by adapting to the varying context (i.e. input data difficulty), Mobiprox can achieve energy savings while preserving the inference accuracy.

\end{itemize}

%% file: tex/related_work.tex
\section{Related Work}
\label{sec:related_work}

\textbf{Resource-efficient deep learning on mobiles.} The expansion of mobile deep learning (DL) applications has been hindered by the high resource consumption of DL models and the difficulty of the edge computing devices, such as battery-powered smartphones, to meet the resource and energy requirements of such applications~\cite{chen2019deep}. Model representations may including hundreds of millions of parameters and  performing the classification of a single input vector can easily overwhelm the available computing and memory resources of edge computing platforms~\cite{LaneSqueezing2017}.

Efforts have, thus, focused on reducing the complexity, while preserving the inference power of DL models through weight quantization~\cite{wu2016quantized}, \blue{pruning~\cite{niu2020patdnn,zhao2023automatic,he2018amc}},  knowledge distillation~\cite{hinton2015distilling} and other methods~\cite{cheng2017survey}. High-level DL frameworks, such as PyTorch, do not readily support mobile platforms, thus, there are relatively few demonstrations of an on-device DL optimization. On the pruning front, PatDNN enables real-time inference using large-scale DL models (e.g., VGG-16, ResNet-50) on mobile devices by harnessing pattern-based model pruning~\cite{niu2020patdnn}, while DeepIoT~\cite{yao2017deepiot} uses reinforcement learning to guide the pruning process. Both solutions lead to significant model size reductions ($90\%$ to $98.9\%$ in case of DeepIoT) and speedups (up to $44.5\times$ in case of PatDNN) with no inference accuracy degradation in certain settings, demonstrating vast opportunities for mobile DL optimisation. Parameter quantization, on the other hand, despite being actively researched~\cite{wu2016quantized, Jin_2020_CVPR,zhou2018adaptive}, sees only limited implementation in the mobile realm. The main reason is the lack of support for arbitrary bit-width computation in today's mobile hardware.

\textbf{Dynamic compression adaptation.} \blue{All of the above approaches share a common drawback: once the approximation is applied, the resulting network remains unchanged during runtime. Thus, such approaches enable operation at a single fixed point on the\textit{ accuracy-resource usage} trade-off curve regardless of how the context in which inference is performed changes during runtime.} However, this operation is inappropriate for the mobile domain, since the changing context of use is a defining trait of mobile computing and can significantly affect the requirements imposed on the DL inference. For instance, a smartphone may or may not be connected to a charger, calling for more or less energy-efficient operation; sensor data may be more or less noisy, requiring more or less complex DL models; depending on the intended use, a user may require more or less accurate inference results from a mobile app. Recent research therefore focuses on enabling accuracy-resource usage trade-off by dynamically adjusting the compression level without the need for re-training the network.

The initial solutions enabling dynamic adaptivity, such as MCDNN~\cite{han2016mcdnn}, relied on having several differently-compressed candidate DL models in the cloud and downloading the most appropriate model on the device according to the current context. While enabling context-adaptation, this strategy adds substantial overheads of model transfer. Early exit networks can dynamically reduce the computational complexity of a single model by not traversing all network layers and halting the computation at one of intermediate exit points in the network instead~\cite{teerapittayanon2016branchynet}. 
SPINN~\cite{laskaridis2020spinn} introduces a scheduler that co-optimises the early-exit policy and DL model splitting at run time, in order to adapt to dynamic conditions and meet user-defined service-level requirements in a cloud-edge environment. The drawbacks of early-exit schemes include the need for off-the-shelf models to be re-structured and re-trained and the complexity of developing exiting policies that will be suitable for a particular operational domain. Unfortunately, despite intended to work in dynamic environments, neither MCDNN nor SPINN have been implemented on mobiles. \blue{DeepX compresses network layers using singular value decomposition and enables execution on heterogeneous mobile hardware. However, it supports only fully connected NN layers and the project code is not publicly available.}


Finally, pruning and quantization have also been revised to support dynamic adaptation. Runtime Neural Pruning (RNP) framework~\cite{10.5555/3294771.3294979} enables bottom-up, layer-by-layer pruning guided by a Markov decision process and reinforcement learning. The importance of each convolutional kernel is assessed and based on it channel-wise pruning is performed, where the network is pruned more when classifying an ``easy-to-classify'' input instance. 
A different approach for dynamic compression adaptation is the Slimmable Neural Network (SNN)~\cite{yu2018slimmable}. The method trains a single CNN and then executes it at different widths (number of channels in a layer), permitting runtime adaptive accuracy-efficiency trade-offs at runtime. There is no need for re-training or loading different models: the network adjusts its width on the fly, based on the resource constraints, input difficulty, or other contextual factors. Any-Precision approach~\cite{yu2020any} proposes a CNN training method that allows the network to adapt its numerical precision during inference to support dynamic speed and accuracy trade-off.
Yet, neither RNP, SNN, nor Any-Precision apply to already trained networks, nor have these techniques been implemented in the mobile realm.  The reason for this is that, unlike our approach, the above methods were not originally planned with mobile platform restrictions in mind. SNNs, for instance, rely on dynamic neural network graph reconfiguration, something that none of the mobile DL frameworks  (e.g. TensorFlow Lite, Pytorch Mobile, etc.) supports at the moment.

%% file: tex/preliminaries.tex
\section{Preliminaries}
\label{sec:preliminaries}

Mobiprox builds upon the existing work on heterogeneous and approximate computing compilers:

\textbf{HPVM} (\textit{Heterogeneous Parallel Virtual Machine})~\cite{kotsifakou2018hpvm} is a compiler infrastructure targeting heterogeneous hardware. It introduces HPVM-C, a programming language for defining \textit{data flow graphs} (DFGs), directed graphs in which nodes represent a computation task and edges represent inputs and outputs of a computation task. Computation workloads are defined using HPVM's intrinsic functions used to specify the {target device} the node will be executed on, node {inputs}, node {outputs}, and any compute {operations} (e.g. addition). HPVM compiler achieves parallel execution of produced binaries by identifying dependencies among the nodes in a DFG and generating compute code for specified target devices (CPU, GPU) for each node. 

\textbf{ApproxHPVM}~\cite{sharif2019approxhpvm} expands HPVM by introducing support for NN tensor operations: multiplication, convolution, addition, pooling, and activation functions.  Additionally, ApproxHPVM enables transforming high level descriptions of convolutional neural networks (in frameworks such as Keras, PyTorch) into DFGs in the form of generated HPVM-C source files. However, while HPVM generates code for computation nodes in a DFG, ApproxHPVM's tensor operations are mapped to functions defined in the \textbf{HPVM Tensor Runtime} library. In ApproxHPVM individual tensor operations can be marked with the maximum allowed level of approximation, and the compiler then ensures that these are mapped to the appropriate underlying approximate computing techniques (either software or hardware-based). Yet, ApproxHPVM's tensor operations are supported for Nvidia CUDA-enabled devices only. 
In this paper we introduce a novel OpenCL implementation that enables approximate tensor operation execution on Android devices (Section~\ref{sec:android-rt}).

\textbf{ApproxTuner}~\cite{sharif2021approxtuner} delivers heuristic-based search of the space of possible approximations of each individual network layer, so that a comprehensive \textit{speedup-inference accuracy} trade-off curve is charted and the list of the most promising sets of approximations is identified. Yet, ApproxTuner does not take into account the peculiarities of the mobile platform and the predicted trade-off curves it draws do not reflect the actual performance observed on the mobiles. Consequently, in this paper we build a new cross-platform approximation profiler based on ApproxTuner (Section~\ref{sec:hpvm-profiler-android}).

\subsection{Approximation techniques}
\label{sec:preliminaries-acts}

\newcommand{\ttit}[1]{\textit{\texttt{#1}}}



\begin{figure} 
    \centering
  \subfloat[Row perforation\label{fig:convperf-row}]{%
       \includegraphics[width=0.45\linewidth]{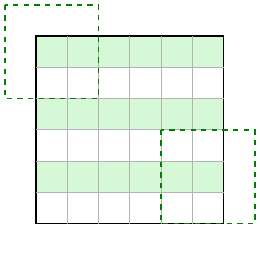}}
    \hfill
  \subfloat[Column perforation\label{fig:convperf-column}]{%
        \includegraphics[width=0.45\linewidth]{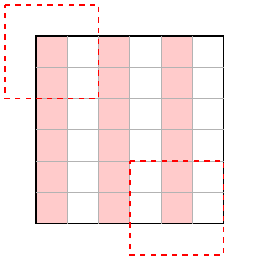}}
    
  \caption{Perforated convolution. Coloured sections indicate convolution coordinates. Dashed squares indicate the area of the first and the final convolution.}
  \label{fig:convperf} 
\end{figure}

We identified the following generally-applicable approximation techniques that can be employed at a level of a single NN operation and are supported by commodity mobile hardware, and we implemented them in Mobiprox:


\textbf{Convolution perforation}~\cite{figurnov2016perforatedcnns} is an approximation that skips certain input matrix coordinates when calculating convolution, as shown in Figure~\ref{fig:convperf}. 
Due to the nature of convolutions, this does not necessarily mean that the inputs at skipped coordinates are never used -- indeed, the inputs are used in neighboring convolutions. This, in turn, makes it feasible to interpolate convolution results at skipped coordinates by computing the average of computed neighboring cells. We support two types of convolution perforation -- \textbf{row perforation} and \textbf{column perforation}. The parameter \ttit{offset} defines the index of the first omitted row or column, while parameter \ttit{stride} defines the interval between the skipped rows/columns.  In Figure~\ref{fig:convperf}, parameters \ttit{stride}=2 and \ttit{offset}=1 were used.



\newcommand{\nelm}{n_\mathrm{elm}}
\newcommand{\nelmPerf}{n_\mathrm{elm-samp}}

\textbf{Filter sampling} approximates the filters that the convolutions are performed with.
In CNNs filters are 4-dimensional tensors with dimensions $[N, C, H, W]$. $N$ represents the number of filters in the convolution, $C$ is the number of feature channels in the input and the filter, and $H$ and $W$ represent the height and width of the filter, respectively. Each filter is therefore composed of $\nelm{} = C \cdot H \cdot W$ components. Filter sampling with \ttit{stride} $k$ removes every $k$-th component of the filter's $\nelm{}$ components, starting at element specified by \ttit{offset}. The technique, thus, reduces the amount of computation by keeping only $\nelmPerf{} = \nelm{} - \frac{\nelm{} - \ttit{offset}}{\ttit{stride}}$ filter components at the cost of the overall convolution accuracy. To interpolate missing values, each retained filter component is multiplied by a factor of $\nelm{}/\nelmPerf{}$.




Finally, Mobiprox also provides an optional \textbf{half-precision quantization} that can be used to approximate any floating point tensor operation. While such quantization is meaningful only if the underlying hardware supports it, we opted for enabling it as modern mobile GPUs, such as those of Arm Mali series, natively support the IEEE FP16 16-bit format.


    


%% file: tex/mobiprox.tex
\section{Mobiprox Framework}
\label{sec:mobiprox}

\begin{figure}[t]
    \centering
    \includegraphics[width=\linewidth]{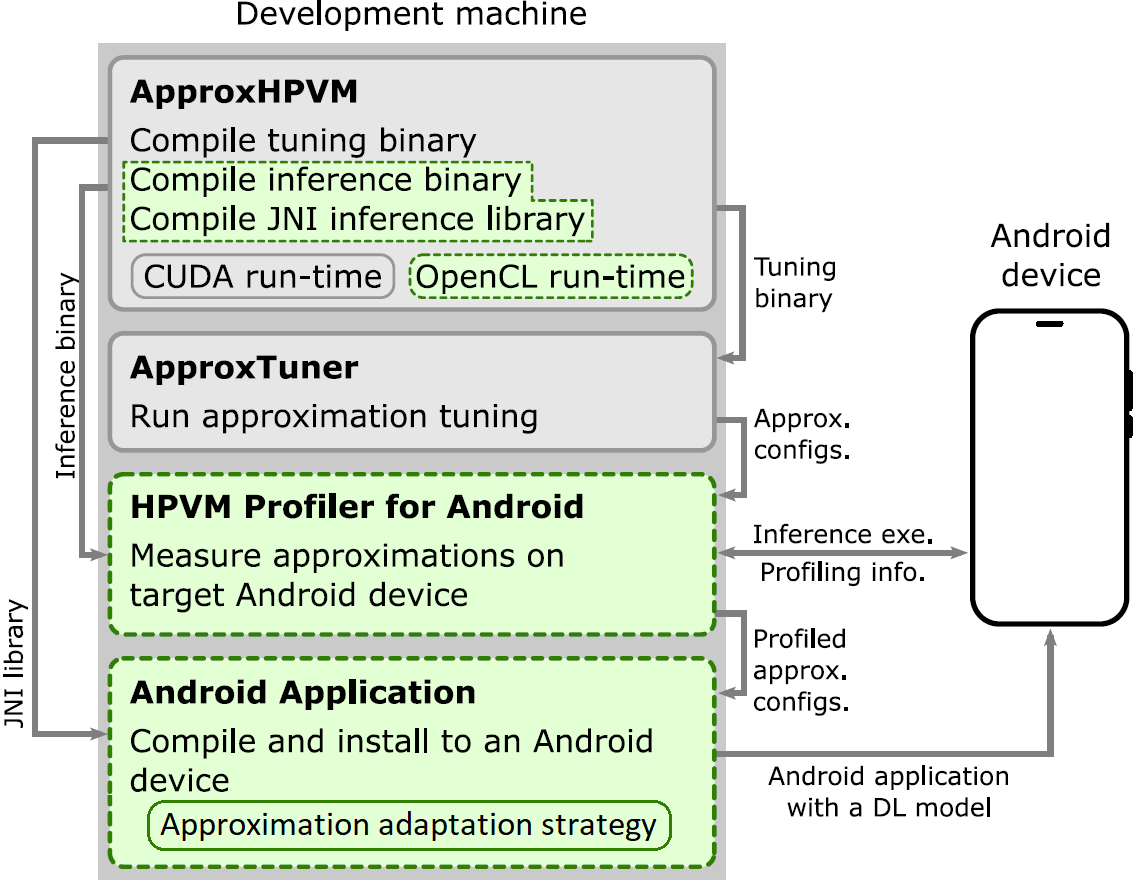}
    \caption{Mobiprox overview. OpenCL run-time supports running the inference binary (controlled either directly from the C code, or via JNI from the main Java/Kotlin app) with a varying level of approximation. The HPVM Profiler for Android helps us chart the \textit{approximation -- resource usage} space, so that the Approximation adaptation strategy wihtin the Android app can set the approximation level dynamically at runtime. Main Mobiprox modules are colored green, while the supporting pre-existing modules are grayed out.}
    \label{fig:amc-pipeline}
\end{figure}

Mobiprox, our novel framework for enabling dynamic approximation of mobile DL, is sketched in Figure~\ref{fig:amc-pipeline}. An Android app compiled with Mobiprox can use an arbitrary runtime approximation adaptation strategy for its DL models (e.g., ``run low quality network when battery is low'', ``run high quality inference when user is at a specific location'', etc.). To achieve this, Mobiprox operates with approximation configurations, i.e. combinations of per-layer approximations of a pre-trained DL model. Mobiprox first uses ApproxTuner to examine the impact of different configurations on the inference accuracy and the speedup. Each approximation configuration yields a point in the accuracy--speedup space, and Mobiprox identifies the most promising configurations that form the Pareto front in this space and then profiles their actual performance on the mobile platform using the novel \textit{HPVM Profilier for Android}. Mobiprox's Android-based \textit{OpenCL runtime} then enables execution of and dynamic switching between approximation configurations on a mobile device. Using the \textit{JNI interface library} generated by the framework, the mobile application can control the approximation level of the NN. Finally, as part of Mobiprox, we also devise \textit{Approximation adaptation strategies }that leverage the generated trade-off curves to match the required and delivered quality of computation, thus enable energy-efficient DL on mobile devices. 

\subsection{Charting approximation space} 
\label{sec:mobiprox:charting}


Each of the approximation techniques described in Section~\ref{sec:preliminaries-acts} exposes one or more \textbf{approximation knobs} that can change the level of approximation and thus adjust the accuracy and the execution time (consequently the energy efficiency) of a tensor operation. These knobs are \ttit{offset} and \ttit{stride} for convolution perforation and filter sampling, and an indicator \texttt{\_fp16} of whether an operation is executed using half-precision quantization. An \textbf{approximation configuration} is a set of pairs $\langle \mathrm{Op.}, \mathrm{KnobValue} \rangle$ for all operations in a given NN. Each of the configurations leads to a single \textbf{trade-off point} on an speedup-accuracy trade-off curve. 

The tuner heuristically searches the space of possible approximations and determines a Pareto frontier of approximation configurations that maximise the execution speedup at different \textbf{quality of service (QoS) loss} points. This loss is a real number defined as a difference between the classification accuracy, over a representative validation dataset, of a non-approximated and an approximated DL model. 

\begin{figure} 
    \centering
  \subfloat[Tuning on server\label{fig:combined-pareto}]{%
       \includegraphics[width=\linewidth]{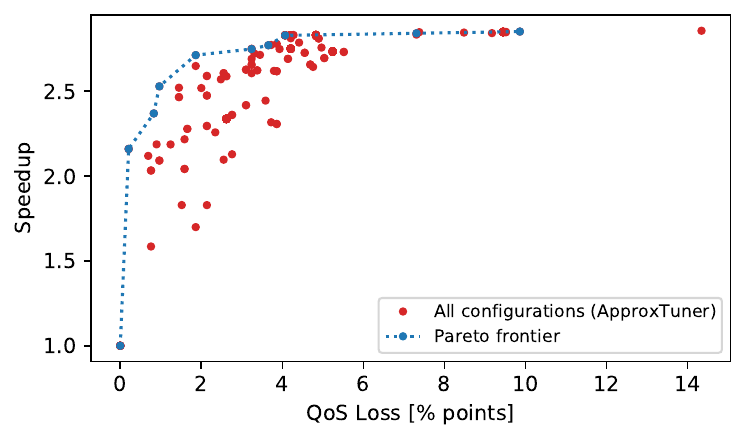}}
    \hfill
  \subfloat[Tuning on mobile\label{fig:combined-pareto-android}]{%
        \includegraphics[width=\linewidth]{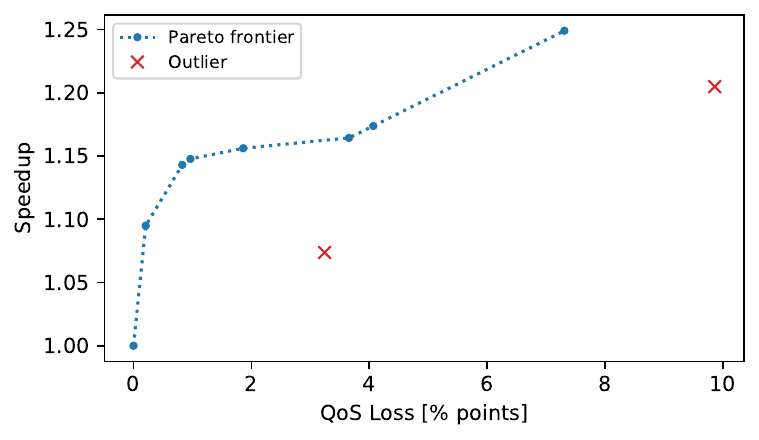}}
    
  \caption{Comparison of the achieved speedup and the resulting QoS (inference accuracy) loss for approximation configurations selected by the on-server tuning with the same configurations ran on a mobile platform. Note the different scaling of the y-axis.}
  \label{fig:pareto} 
\end{figure}

However, the method described above for determining the optimal approximation configurations does not readily translate to mobile devices. The mobile platform is substantially different from the server used for fast heuristic-based approximation configuration profiling. The specifics of GPU-based execution (e.g., CUDA vs OpenCL), heterogeneous CPUs with fewer cores, and other factors mean that the results of the configuration search performed on a server are a rather poor representation of the actual approximated NN performance on a mobile. In Figure~\ref{fig:combined-pareto-android}, on the example of a MobileNetV2 model used for HAR (detailed in Section~\ref{sec:methodology}), we show the actual on-mobile-device speedup and QoS loss achieved by the approximation configurations that ApproxTuner identified as the most promising. While the Pareto points obtained on a server generally remain relevant, the achieved on-device speedup is about 50\% lower on the mobile.


Mobiprox therefore introduces a novel configuration identification approach. First, we perform tuning on a computer cluster to identify candidate approximation configurations. Then, we develop an Android-based profiler (described in Section~\ref{sec:hpvm-profiler-android}) that runs each candidate configuration on a mobile device and obtains a realistic picture of the approximated neural network performance. The resulting picture of the speedup -- QoS loss space charted by these configurations is then used to guide the dynamic adaptation of the approximation. As a final result, the profiler creates a file listing configurations that will be switched during the mobile app runtime (according to a strategy, e.g. from Section~\ref{sec:strategies}), yet\textit{ only a single network model definition gets deployed on a mobile}.


\subsection{Mobiprox -- Android implementation}

Mobiprox, as a concept, is not tied to a particular mobile platform. Yet, amassing 75\% of the smartphone market share Android is the most common mobile deep learning platform and that stands to gain the most from dynamically adaptable approximation, thus, in this section we develop a full Mobiprox compilation pipeline targeting Android devices. 

\subsubsection{Mobiprox Android Compiler}
\label{sec:ndk-integration}

\newcommand{\ndkversion}{21.4.7075529}

Mobile application development with Mobiprox involves compiling the tuning binary and the inference binary (Figure~\ref{fig:amc-pipeline}). While the tuning binary is confined to the server environment and is handled by the ApproxHPVM compilation pipeline, the inference binary is cross-compiled from a server to a mobile (Android). We implement a mechanism for turn-taking between ApproxHPVM and Android NDK LLVM compiler toolchains (Algorithm~\ref{alg:llvm-compile}). We enable this by clearly partitioning the compilation steps and harnessing the fact that LLVM-based compilers apply transformations to an intermediate representation termed LLVM-IR. Note that ApproxHPVM extends LLVM-IR by defining HPVM-IR to which approximation-related transformations are applied. This clear division allows us to use Android NDK for generating the initial LLVM-IR suitable for Android applications and for generating the machine code containing approximate NN operations suitable for mobile GPUs in the final compilation step, while using HPVM-IR transformations for the internal part of the compilation pipeline to insert the description of the desired approximate tensor operations.

\begin{algorithm}[h]
\SetAlgoLined
\caption{Mobiprox compilation. Compilers used at each step are shown in comments.}
\label{alg:llvm-compile}
\newcommand{\IRL}{IR_{LLVM}}
\newcommand{\IRH}{{IR}_{HPVM}}
\begin{scriptsize}
   
$\IRL \gets$ Transform source code into LLVM-IR \tcp*{\textbf{Android LLVM}}
$\IRH \gets$ Transform $\IRL$ into HPVM-IR \tcp*{\textbf{ApproxHPVM}}
\For{\textbf{each} IR transformation $T_i$ of the compiler} {
$\IRH \gets T_i(\IRH)$ \tcp*{\textbf{ApproxHPVM}}
}
$\IRL \gets$ Transform $\IRH$ into LLVM-IR \tcp*{\textbf{ApproxHPVM}}
Compile $\IRL$ to machine code \tcp*{\textbf{Android LLVM}}
\end{scriptsize}
\end{algorithm}

\subsubsection{OpenCL Tensor Runtime for Android}
\label{sec:android-rt}

A core component of Mobiprox Android is a Tensor Runtime, which implements tuneable approximable tensor operations for NN inference. The existing support for approximate NN operations for Nvidia CUDA GPUs~\cite{sharif2019approxhpvm} is not suitable for mobiles, which seldom host such hardware. Instead, Mobiprox implements an own tensor runtime using OpenCL, an open standard for GPU-accelerated computation which is available on a wide variety of hardware, including mobile platforms.

%
    
%

To enable an enhanced control over low-level concepts (such as memory allocation), we implemented the tensor runtime for Android using CLBlast~\cite{clblast}, an OpenCL implementation of basic linear algebra subprograms (BLAS). However, this library is not intended for deep learning: it does not implement operations commonly used in NNs. Therefore we extended CLBlast with the following operators: \textit{i)} Point-wise tensor addition, \textit{ii)} Bias addition, \textit{iii)} Activation functions (ReLU, clipped ReLU, $tanh$), \textit{iv)} FP-16 -- FP-32 tensor conversion, \textit{v)} Batch normalisation, \textit{vi)} Pooling ($min$, $max$, $average$), \textit{vii)} Convolution approximations operators optimized with tiling and vectorization: \textit{Image-to-Column} ($im2col$) transformations with row perforation, column perforation, and filter sampling, \textit{Kernel-to-Row} ($kn2row$) transformation with filter sampling, and \textit{Interpolation} of missing values in convolution perforation. 
%
Finally, during the mobile app compilation Java Native Interface (JNI) is exposed, enabling the tensor runtime initialization and destruction, NN inference invocation, and dynamic approximation configuration loading.

\subsubsection{HPVM Profiler for Android}
\label{sec:hpvm-profiler-android}

To assess the speedups and consequently the energy efficiency of approximated NNs we implement a profiler tool. The profiler, in the form of a Python library, for a given NN binary measures the accuracy, softmax confidence, and execution time of NN inference on a given test dataset. Due to a high discrepancy between the speedup observed on a mobile device and on a server for the same approximated network (Figure~\ref{fig:pareto}), the profiler uses the Android Debug Bridge (ADB)~\cite{adb} to run measurements on an actual Android device and to transfer the profiling information files back to the host machine for analysis.

%% file: tex/strategies.tex
\section{Approximation adaptation strategies}
\label{sec:strategies}



Mobiprox's key strength is its support for context-based adaptation of mobile DL approximation. The framework itself deliberately does not prescribe the adaptation strategy allowing a developer to implement an arbitrary set of rules driven by energy needs (e.g. ``use higher approximation when battery level falls below 10\%''), the purpose of use (e.g. ``use more accurate HAR models when a user is exercising''), or even business models (e.g. ``use input-adaptable approximation for premium users''). 
Programming such strategies is trivial, yet, one can envision a more challenging-to-achieve goal, such as ``minimize the energy usage without sacrificing the inference accuracy''. In this section we harness the natural temporal dependence of the instances of sensed data that is characteristic in many mobile computing applications, and devise two strategies demonstrating that a widely applicable goal of energy minimization can be met with Mobiprox.

\subsection{State-driven}

Many mobile sensing domains deal with the recognition of states that do not vary rapidly over time: human physiological signals do not change erratically, people have conversations, not random utterances, movement is continuous in space, etc. Our state-driven adaptation strategy is based on the observation that rapid variations, especially in human behavior, are rare (e.g.~\cite{jabla2019balancing}). We hypothesise that inputs that are less difficult to classify can be processed with more ``aggresive'' energy-saving approximation configurations, whereas more difficult-to-classify inputs require computationally more expensive, more accurate configurations, and that the ``difficulty'' of the input correlates with the class an instance belongs to. 




Starting from this assumptions we implement an adaptation algorithm that adjusts the approximation configuration based on the reliability of classification determined by looking at a subset of the most recent predictions made by the network. After each inference, a vote is cast on the measure of reliability $V$, which is increased by $1$ if all previous $N$ predictions are equal, and decreased by $1$ otherwise. The functionality of this approach is described in detail in Algorithm~\ref{alg:adapt-SM}. 

In this algorithm, $V_{L}$ refers to the number of required votes that need to be cast consecutively in order to change the approximation configuration -- this parameter avoids the situation where the configuration is changed at every inference point. 
The second parameter $N$ defines the capacity of the FIFO memory $M$. A larger memory would increase the robustness of the algorithm to classification errors (since it will consider a larger subset of previous predictions), but at the same time would hinder switching to more approximate configurations after a change in the observed/modeled phenomenon.


    
    

\begin{algorithm}[!htb]
\caption{State-driven adaptation engine}
\label{alg:adapt-SM}
\begin{scriptsize}
\SetKwInOut{Input}{Input}
\SetKwInOut{Output}{Output}


$M = []$ \tcp*{FIFO memory with maximal capacity $N$}
$V = 0$ \tcp*{Reliability index on interval $[-V_L, V_L]$}

\While{$p$ = nextPrediction()}{
$\mathrm{push}(M, p)$;

\If{$\mathrm{len}(M) < N$}{
continue;
}

\If{all predictions in $M$ are equal}{
    $V = \mathrm{max}(0, V) + 1$\;
}
\Else{
    $V = \mathrm{min}(0, V) - 1$\;
}

\If{$V \leq -V_L$}{
    Approximate \textit{less}\;
}
\ElseIf{$V \geq V_L$}{
    Approximate \textit{more}\;
}
pop(M);
}
\end{scriptsize}
\end{algorithm}
\vspace{-9pt}
\subsection{Confidence-driven}



In the second adaptation strategy, we use the classifier's confidence as a proxy for accuracy. The softmax layer probability can accurately reflect the actual confidence of the classifier~\cite{mahmoud2021optimizing}. However, Guo et al.~\cite{guo2017calibration} point out that calibration is required to achieve a high correlation between the softmax confidence and the expected inference accuracy. Hence, we perform calibration by applying the temperature scaling during softmax confidence calculation. More specifically, for an $N$-class classification task where the $N$-dimensional vector $z$ contains class scores, for any class $i$, its calibrated softmax confidence is computed as:
 
\begin{equation}
    \sigma_i(z; T) = \frac{e^{z_i/T}}{\sum_{j=1}^{N} e^{z_j/T}}
\end{equation}
 
\noindent where $T$ is a scalar temperature parameter, which ``softens'' the softmax (raises the output entropy) when $T>1$ and is optimized with respect to negative log likelihood on the validation dataset, so that the confidence value for the datapoints classified with accuracy $p$ is as close a possible to $p$~\cite{guo2017calibration}. 

\newcommand{\cCorrect}{C_{\mathrm{+}}^{(i)}}
\newcommand{\cWrong}{C_{\mathrm{-}}^{(i)}}
\newcommand{\tMore}{T_{\mathrm{more}}}
\newcommand{\tLess}{T_{\mathrm{less}}}
\newcommand{\CMore}{C_{\mathrm{more}}^{(i)}}
\newcommand{\CLess}{C_{\mathrm{less}}^{(i)}}

Our adaptation strategy then uses the calibrated softmax confidence to identify incorrect classifications. The Android profiler (Section~\ref{sec:hpvm-profiler-android}) also reports per-class confidence averages for correct ($\cCorrect{}$) and incorrect ($\cWrong{}$) predictions and adds this information to approximation configuration files. The algorithm is then driven by a hysteresis outlined by two thresholds $\CLess$ and $\CMore$, where $\cWrong{}>\CLess>\CMore>\cCorrect{}$. If the classification confidence of the predicted class of the immediately preceding instance is higher than $\CMore$, the algorithm moves towards more aggressive approximation. If it is lower than $\CLess$, the algorithm moves towards less approximated configuration. We empirically find that the values of $\CLess$ halfway and $\CMore$ three-quarters-way between $\cWrong{}$ and $\cCorrect{}$, respectively, perform well in our experiments.

    
    

%% file: tex/methodology.tex
\section{Experimental Setup}
\label{sec:methodology}

\blue{To evaluate our framework we first, through a series of microbenchmarks, assess the energy savings and speedup achieved through approximate neural networking operations implemented in Mobiprox, and also evaluate the overhead incurred by the accuracy--speedup profiling that Mobiprox relies on. Then, we evaluate Mobiprox's dynamic adaptation in two domains -- human activity recognition and spoken keyword recognition. Finally, we deploy Mobiprox on commodity Android phones and demonstrate its usability for real-time adaptable DL inference. }

\newcommand{\MPM}{Monsoon Power Monitor}
\newcommand{\tboard}{ASUS TinkerBoard S}

\blue{\textbf{Microbenchmarks with standard architectures.}} 
We evaluate Mobiprox first through a series of experiments aiming to assess the energy savings and speedup achieved through different approximate NN operations over a selection of networks. \blue{We first investigate the performance on standard image recognition architectures (AlexNet, VGG16, and MobileNet) and the CIFAR-10 dataset. However, since Mobiprox primarily targets dynamic mobile environments and inference from time-series sensor data, we also include two NN architectures (MobileNet and ResNet50) trained on UCI-HAR human activity recognition dataset~\cite{anguita2013public}. We implement all networks in PyTorch.}

Mobiprox is fully compliant with consumer off-the-shelf Android devices. Yet, modern unibody smartphones do not allow for batteries to be easily removed, precluding the use of high-accuracy power metering. Therefore, when energy consumption is examined, we use 
{\tboard}\footnote{\url{https://tinker-board.asus.com/product/tinker-board-s.html}} \blue{single-board computer} running Android 7 OS. We power \blue{it} through the high-frequency \MPM{} and use the accompanying PowerTool\footnote{ \url{https://www.msoon.com/hvpm-software-download}} measurement processing software. Our Python-based profiler using ADB runs compiled approximated NNs on the board. The approximation's main, yet not the only (as we will see in Section~\ref{sec:evaluation_microbenchmarks}) impact on the energy consumption stems from the decreased DL processing time.  The profiling for each network is, thus, executed on a predefined fraction of the data in 10 batches for UCI-HAR networks and in 8 batches for CIFAR-10. This was done to 
    \textit{i)} reduce overall time requirement,
    and
    \textit{ii)} obtain more robust measurements by measuring each batch separately.
We report the mean and standard deviation of each configuration's energy consumption.

\textbf{Real-world human activity recognition traces.} We assess the expected energy savings Mobiprox brings in real-world environments by taking a recent trace of human activity obtained through a body-mounted mobile sensing platform~\cite{electronics10232958}. The dataset contains  traces of 21 participants (13m, 8f), with an average age of 29 (std. dev 12) years. The traces consist of the acceleration and angular velocity in all three axes sampled at 50 Hz from an UDOO Neo Full board\footnote{\url{https://shop.udoo.org/en/udoo-neo-full.html}}, a compact IoT embedded computing device equipped with an accelerometer and a digital gyroscope, strapped to each participant's waist. 

In this study, which took place at a university campus, the participants performed the six activities in a row. First the static ones -- sitting, standing still, and lying -- for 2 minutes each. Then, the dynamic activities -- walking up and down a hallway (summing up to two to three minutes for each participant) and walking down and up the stairs (about 45 seconds in each direction, the duration being limited by the total number of stairs). \blue{This experiment features the same six activities that are present in the original UCI-HAR dataset. Yet, by using traces collected with a different device, in a different environment, and with different participants than the original experiment, we aim to obtain a realistic picture of Mobiprox's ability to adapt to previously unseen users. Separately, we conduct an experiment with Mobiprox running directly on Android smartphones that further assesses the performance of our framework with human activity recognition in unscripted scenarios (ellaborated below).}


\textbf{Real-world spoken keyword recognition traces.} We also examine the savings Mobiprox brings in real-world environments by considering the problem of a spoken keyword recognition from microphone recordings. For this we use the Google Speech Commands (GSC) v0.01~\cite{warden2018speech}, a dataset containing 65,000 one-second long utterances of 30 short words by thousands of different speakers. Interested in the recognition of keywords in realistic situations, where a word has to be spotted in sound segments that may also contain words we are not interested in as well as recordings of silence, we follow the approach presented in ~\cite{tang2017honk} and use twelve classes for ten selected keywords (yes, no, up, down, left, right, on, off, stop and go) and two extra classes: ``unknown'' (for the remaining 20 words in the dataset) and ``silence''. In Section~\ref{sec:evaluation} we evaluate Mobiprox's ability to bring energy savings when a compact NN is used for on-device spoken keyword recognition within the GSC-based trace.


\blue{
\textbf{Live smartphone-based human activity recognition.} We recruit ten users (all students or staff at University of Ljubljana, 7 female/3 male) to perform a 10-minute experiment during which they are given an option of conducting six activities -- sitting, standing still, lying, walking, going up the stairs and going down the stairs. Unlike with the lab-based studies, such as~\cite{anguita2013public}, the order and the duration in which the activites were to be performed, was not in any way prescribed in our experiment. A Samsung Galaxy M21 smartphone, in the portrait orientation with the screen facing forward, was attached to each user's waist. The phone sampled accelerometer and gyroscope at 50Hz, and ran Mobiprox for live on-device inference of human activity using dynamically approximated neural network. The model used was \mnUci{} pre-trained on UCI-HAR dataset without any further re-training. Finally, accelerometer and gyroscope samples were stored and re-ran through a non-approximated \mnUci{} model to obtain the baseline activity prediction ({\blue{the output of the non-approximated model is not the ground truth, as due to the unscripted nature of the experiment we do not know the exact activity a user performed at a given moment}}). 


}

%% file: tex/evaluation.tex
\section{Evaluation}
\label{sec:evaluation}

In our evaluation of the Mobiprox framework we aim answer to the following research questions:
\begin{itemize}
    \item \emph{RQ1:} Usability: time to find configurations and generalizability across different devices?
    \item \emph{RQ2:} Generalizability across NN models: how does Mobiprox perform across different neural network models and classification tasks, in terms of accuracy, speedup, and energy saved?
    \item \emph{RQ3:} What energy savings would Mobiprox bring in a real world scenario and at what trade-off with regard to inference accuracy?
\end{itemize}
    
\subsection{Configuration identification time and generalizability}

In Mobiprox, the identification of suitable approximation configurations is split into two phases: \textit{i)} identifying the candidate Pareto front among all possible configurations and \textit{ii)} measuring each configuration's speedup and inference accuracy on the target platform. The first part of the configuration identification process relies on ApproxTuner and is performed on a CUDA GPU-enabled machine. The second part of the process must be executed on a machine (e.g. a PC or a laptop) that connects to the target mobile platform via Android Debug Bridge (ADB). In this step, the candidate configurations get executed directly on the mobile, thus, what matters are the capabilities of the connected mobile platform, not the machine that controls the execution.

In our experiments we use a single node of a grid supercomputer equipped with Nvidia Quadro GV100 GPUs for the first phase of the configuration identification process. The node uses a single GPU for the tuning task, which is executed in a batch processing mode. In Table~\ref{tab:profiling_time_server} we list the times needed for finding the Pareto front of the configurations for different networks used in our experiments. The heuristic search done by ApproxTuner is not deterministic, thus different runs may be completed in slightly different amounts of time. In addition, subsequent tunings of the same network often take significantly less time, as they build upon already cached results. In any case, we observe that the even the most complex tuning (for resnet50\_uci-har) completes in less than 30 minutes. 

\begin{table}[!htbp]
    \caption{Time needed for identifying Pareto-optimal configurations using a single GPU on a supercomputer. Standard deviations are in the parentheses. The duration may vary with different search parameters}
    \begin{center}
    \begin{tabular}{lll}
DL model  & Initial tuning time [s]  & Subsequent tuning time [s]\\ \hline
alexnet2\_cifar & 574 (97)  & 184 (6)\\ 
mobilenet\_cifar10 & 1301 (226) & 370 (18)\\ 
vgg16\_cifar10 & 1216 (62) & 264 (2) \\
resnet50\_uci-har & 1602 (316) & 276 (14)\\ 
mobilenet\_uci-har & 1203 (246) & 253 (1)\\ 
\end{tabular}
    \end{center}
\label{tab:profiling_time_server}
\end{table}

For the second phase of configuration identification we use an ASUS TinkerBoard S, which, with its Rockchip RK3288 system-on-chip released in 2014, represents a lower-end platform. In Table~\ref{tab:profiling_time_client} we list the times needed for measuring the classification accuracy and the speedup for all Pareto front configurations found in the first phase of the identification process. The size of the test dataset, the batch size, as well as the number of configurations in the Pareto front varied for different networks. Nevertheless, this one-off process completes in less than 58 mins. even on our low-end mobile platform, thus, we conclude that Mobiprox can be comfortably used within the existing Android application compilation process. 

\begin{table}[!htbp]
    \caption{Time needed for profiling Pareto-optimal configurations' speedup and accuracy on a low-end mobile device.}
    \begin{center}
    \begin{tabular}{lllll}

DL model  & Dataset (batch) & Configs. & Profiling time [mins] \\ \hline
alexnet2\_cifar &  800 (100) & 20 & 36 \\ 
mobilenet\_cifar10 & 800 (100) & 20 & 58 \\ 
vgg16\_cifar10 & 200 (25) & 20 & 42 \\ 
resnet50\_uci-har & 250 (50) & 20 & 26 \\ 
mobilenet\_uci-har & 1450 (145) & 10 & 42 \\ 
\end{tabular}
    \end{center}
\label{tab:profiling_time_client}
\end{table}

To answer RQ1, we now discuss the generalizability of an application compiled with Mobiprox. Mobiprox requires an actual mobile hardware for the second part of the configuration identification. Tens of thousands of different Android devices exist on the market, however, the choice of which to use should not have a major impact on the approximate configuration profiling. First, despite different hardware, the platforms use the same OpenCL-based primitives we developed in Section~\ref{sec:android-rt}. Second, while the speed at which NN will be executed may differ among different devices, there is no reason to expect that speedups (relative to a non-approximated network) will be different for the same configuration ran on different devices. This is especially true if these devices belong to the same architectural category. Currently, Android apps can be built for four such categories, i.e. Application Binary Interfaces (ABIs). Yet, 99\% of the smartphones rely on one of the two ARM-based ABIs\footnote{\url{https://stackoverflow.com/questions/46453457/which-android-abis-cpu-architectures-do-i-need-to-serve}}, armeabi-v7a and arm64-v8a, both of which we successfully tested on ASUS TinkerBoard S and a range of phones (Samsung Galaxy M21, Samsung Galaxy S21, Xiaomi Pocophone) in our lab. 
For the deep learning models presented in this paper we have not observed any differences in ordering among speedups obtained by different configurations on the platforms we have experimented with. 

\subsection{Popular CNN benchmarks}
\label{sec:evaluation_microbenchmarks}

\begin{figure} 
    \centering
  \subfloat[CIFAR-10\label{fig:energy-cifar}]{%
       \includegraphics[width=\linewidth]{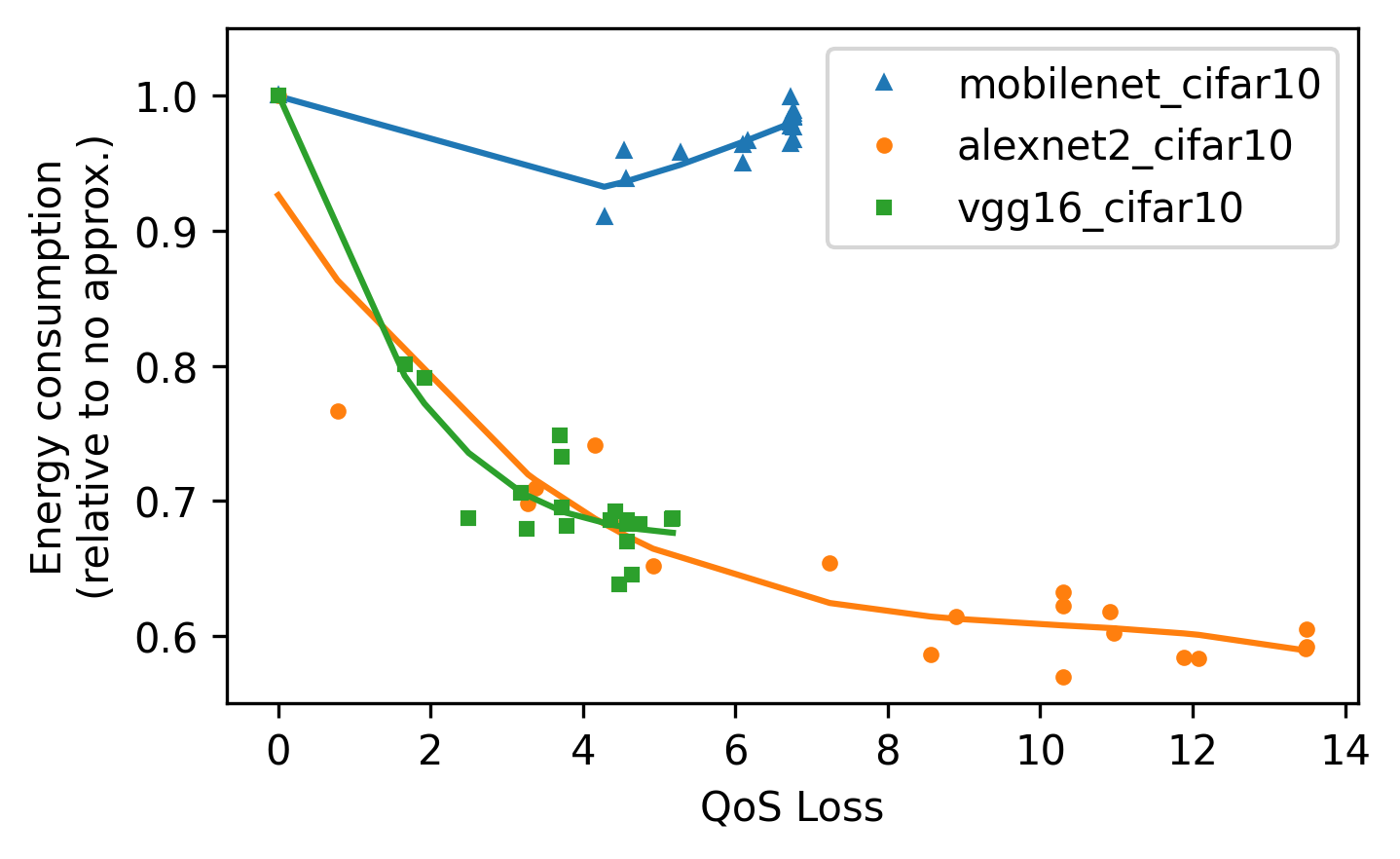}}
    \hfill
  \subfloat[UCI-HAR\label{fig:energy-uci}]{%
        \includegraphics[width=\linewidth]{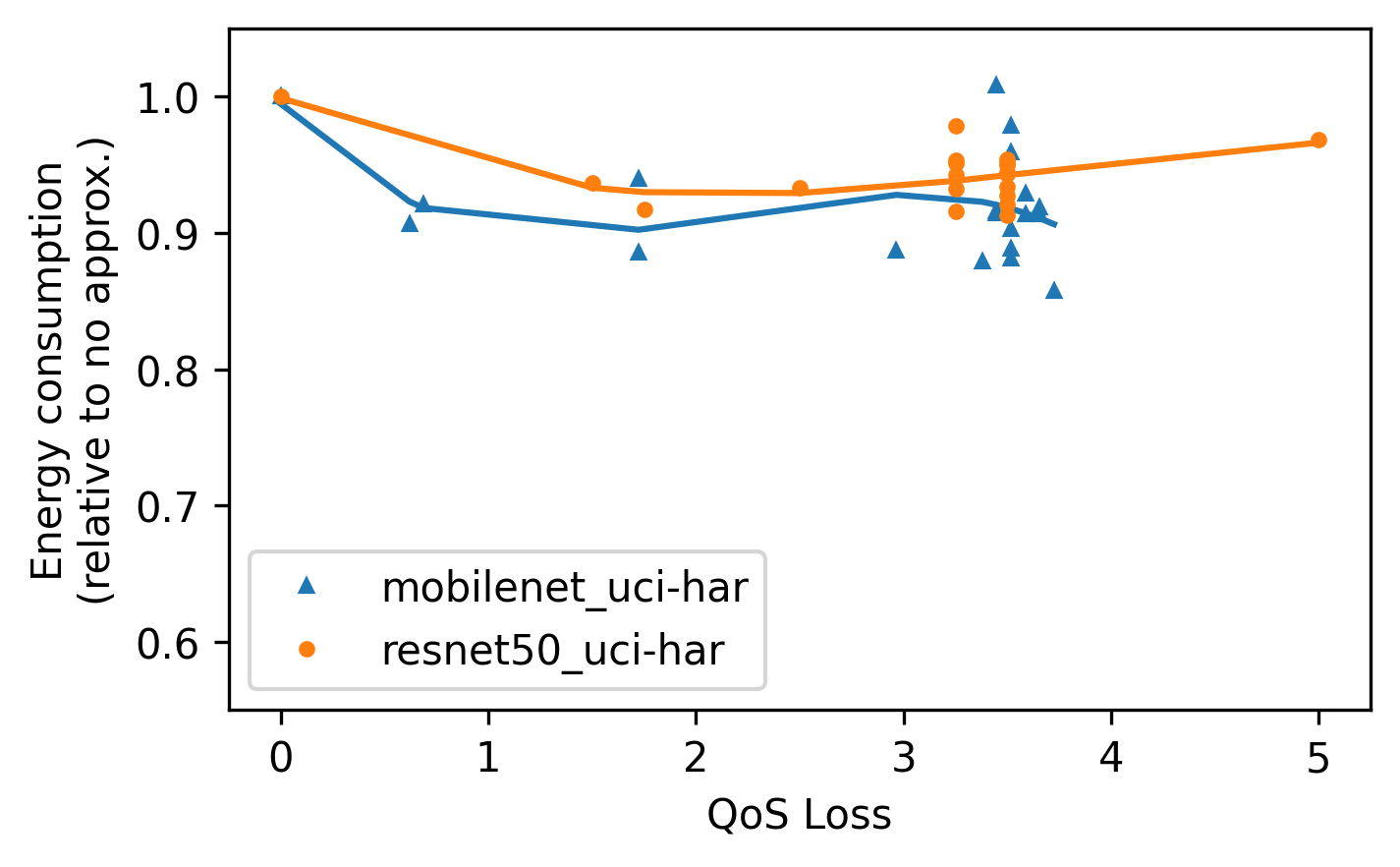}}
    
    \caption{System-wide energy consumption (relative to no approximation) of an \tboard{} running inference on NNs trained different datasets. Different point types correspond to different NN architectures; each point represents a single approx. configuration. The x-axis represents the actual QoS loss from the model deployed on a mobile device.}
    \label{fig:energy} 
\end{figure}

To answer \emph{RQ2}, we obtain approximation configuration sets for NNs trained on CIFAR-10 and UCI-HAR datasets using the Mobiprox tuner\footnote{For clarity, we limit the number of configurations and instruct the tuner not to consider configurations that result in QoS loss below a given threshold.} and present the profiling results in Figures~\ref{fig:energy-cifar} and~\ref{fig:energy-uci}. The reported energy consumption reduction is \textit{system-level}, i.e., idle consumption has not been subtracted. We can observe a key difference between the approximation configurations of different NN architectures -- larger networks, such as VGG and AlexNet, are more amenable to approximation, and Mobiprox yields higher energy savings for these networks. This may suggest that certain models that are prohibitively expensive for mobile DL can be made mobile-ready using Mobiprox even without retraining.

It is interesting to juxtapose the measured energy savings with the speedup expected at the tuning time. In Figure~\ref{fig:combined-pareto} in Section~\ref{sec:mobiprox:charting}, we show that server-based profiling indicates that our approximations can lead to more than $ 2.5 \times$ speedup of inference on the MobileNet architecture trained on the UCI-HAR dataset. The actual reduction in energy consumption is much smaller and is consistent with the speedup measurements presented in Figure~\ref{fig:combined-pareto-android}. We believe that realising the full potential of the approximation on the mobile platform requires careful consideration of the mobile processing hardware. The overheads and thread scheduling inefficiencies in the ARM Cortex-A17 computing architecture on which the experiments were performed may be a likely culprit~\cite{wang2021asymo}.

\begin{figure}
    \centering
    \includegraphics[width=\linewidth]{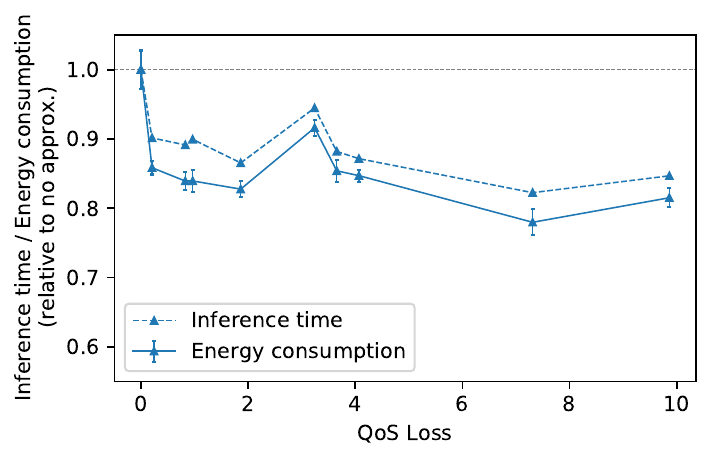}
    \caption{Relative energy consumption compared to relative inference time reduction for \mnUci{} at various approximation configurations. The x-axis shows the actual QoS loss from the model deployed on a mobile.}
    \label{fig:energy-eff-uci}
\end{figure}

In Figure~\ref{fig:energy-eff-uci} we analyse how speedups translate to energy consumption reduction. \blue{We observe a clear relationship between the two lines indicating that a higher speedup indeed reduces energy consumption. Due to dynamic voltage-frequency scaling this relationship does not necessarily hold for any general computing task, as at light loads the CPU/GPU governor might lower the CPU/GPU frequency, and thus, the power consumption. This would lead to a more complex relationship among power, energy, and speedup. However, DL computation is, based on our experience with mobile devices, highly demanding leaving no space for the governor to reduce the frequency and make the speedup -- energy consumption relationship non-trivial. }



The adaptation strategy calculation, i.e. deciding which approximation level to use, plays virtually no role in the overall energy consumption. Irrespective of which of the two strategies presented in Section~\ref{sec:strategies} we employ, the process boils down to either assessing the equality of $M$ predictions, where $M$ is a small integer, and comparing the updated integer reliability metric $V$ with a constant threshold  (``state-driven'' strategy), or comparing the softmax confidence with a constant threshold (``confidence-driven'' strategy). Each of these calculations is performed in a constant time that is negligible compared to an execution time of even a single NN operation. The choice of the strategy, however, impacts the levels that the 
NN will be approximated with. Thus, in the remainder of the evaluation we examine the mobile DL accuracy and energy efficiency afforded by different approximation adaptation strategies.

\subsection{Adaptation strategy evaluation}

To answer \emph{RQ3}, we assess the Mobiprox's ability to deliver energy savings when adaptation according to the strategies developed in Section~\ref{sec:strategies} is performed in realistic dynamic scenarios. This we evaluate in two mobile DL domains:


\subsubsection{Human activity recognition (HAR)}


We run the MobileNet-V2 dynamically approximable NN on the HAR traces described in Section~\ref{sec:methodology}. These traces were collected in a completely different session and by different authors than the original UCI-HAR traces used for the network training. We evaluated the adaptation strategies from Section~\ref{sec:strategies} and compared the results with the ones obtained by the non-approximated MobileNet-V2 (Table~\ref{tab:all_engines}). For each strategy we choose the option for moving to more aggressive approximations (linear vs. exponential) that yielded the best results. 

\begin{table}[!htbp]
    \caption{Inference accuracy and energy consumption on the HAR traces from~\cite{electronics10232958} for MobileNet-V2 trained on the UCI-HAR dataset.}
    \begin{center}
    \begin{tabular}{llll}

Adaptation  & Incr. & Accuracy & Relative Energy \\ \hline
Non-approximated & - & 0.65     & 1.0        \\ 
Confidence-based    & Expon.   & 0.63     & 0.854        \\ 
State-based ($V_{L}=2$)         & Linear   & 0.63     & 0.867        \\ 
\end{tabular}
    \end{center}
\label{tab:all_engines}
\end{table}

These results show that all adaptation engines are more energy efficient than the vanilla MobileNet-V2 with a small drop in average accuracy. The optimal trade-off between the energy saved and the drop in accuracy is obtained using the Confidence-based adaptation engine, which is 15\% more energy efficient with just a 2\% drop in overall average accuracy. The accuracy results are modest (the accuracy of the non-approximated network on the UCI-HAR test set was 90\%), which is to be expected given that we used a network trained on a dataset collected in one environment for performing human activity inference on data collected in a completely different environment. Thus, the results are more in line with other efforts involving HAR on free-living data for using networks trained on the UCI-HAR dataset~\cite{cruciani2020feature}.

\begin{figure}[!htbp]
    \centering
    \includegraphics[width=0.9\linewidth]{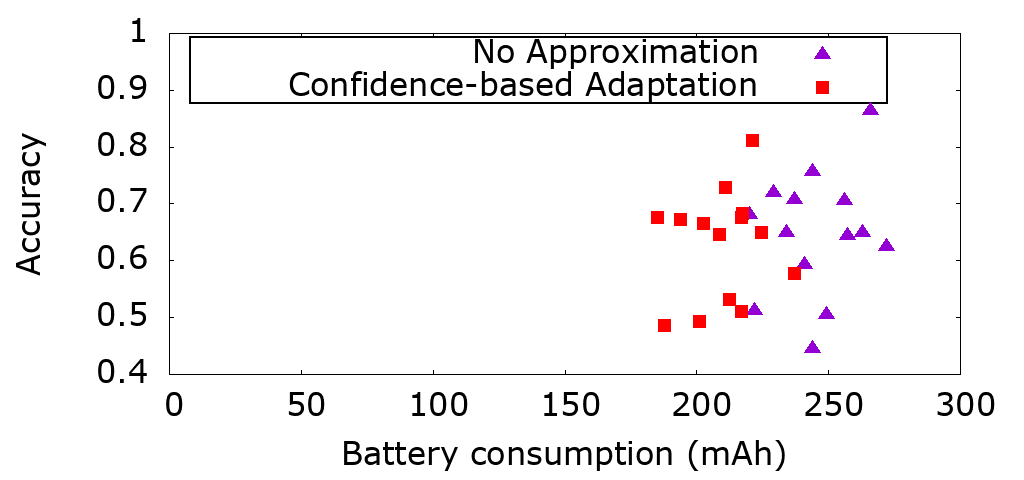}
    \caption{Average accuracy vs. average energy consumption for each user for the non-approximated network and the confidence-based adaptation.}
    \label{fig:acc-vs-energy}
\end{figure}

Finally, to understand whether the accuracy-energy impact is uniform across the users, in Figure~\ref{fig:acc-vs-energy} we show the average accuracy vs. average energy consumption for each user trace for both the non-approximated network and the approximated network using the confidence-based adaptation engine. There is a general trend in the reduction of the energy consumption while maintaining the comparable classification accuracy.

\subsubsection{Spoken keyword recognition (SKR)}  

Mobiprox approximation strategies are not restricted to a particular domain. Thus, we also demonstrate approximation adaptation of a SKR DL model. Understanding voice commands is a critical affordance in many ubiquitous computing settings, such as, for providing driving assistance, or smart home functionalities. 

From a number of DL models have been crafted for SKR we focus on CNN-based models introduced by Sainath and Parada~\cite{sainath2015convspeech}. The family of models presented in this work is light-weight, both in terms of memory usage and computation requirements, thus, well-suited for mobile devices. We adopt a particular PyTorch implementation of a model from this family consisting of two convolutional layers and one fully-connected layer with the softmax output presented in~\cite{tang2017honk}. The code accompanying~\cite{tang2017honk} already contains the DL model weights obtained through training on the 80\% of the GSC dataset (validation on 10\%), and we reuse these weights in our model. This model is then funneled to Mobiprox's on-server and later on-device tuning on the ASUS Tinkerboard S to obtain a 10-point Pareto front of approximate configurations of the network. For tuning we use a half of the 10\% of the GSC that was not used for the training/validation.



\begin{figure}[!htbp]
    \centering
    \includegraphics[width=0.9\linewidth]{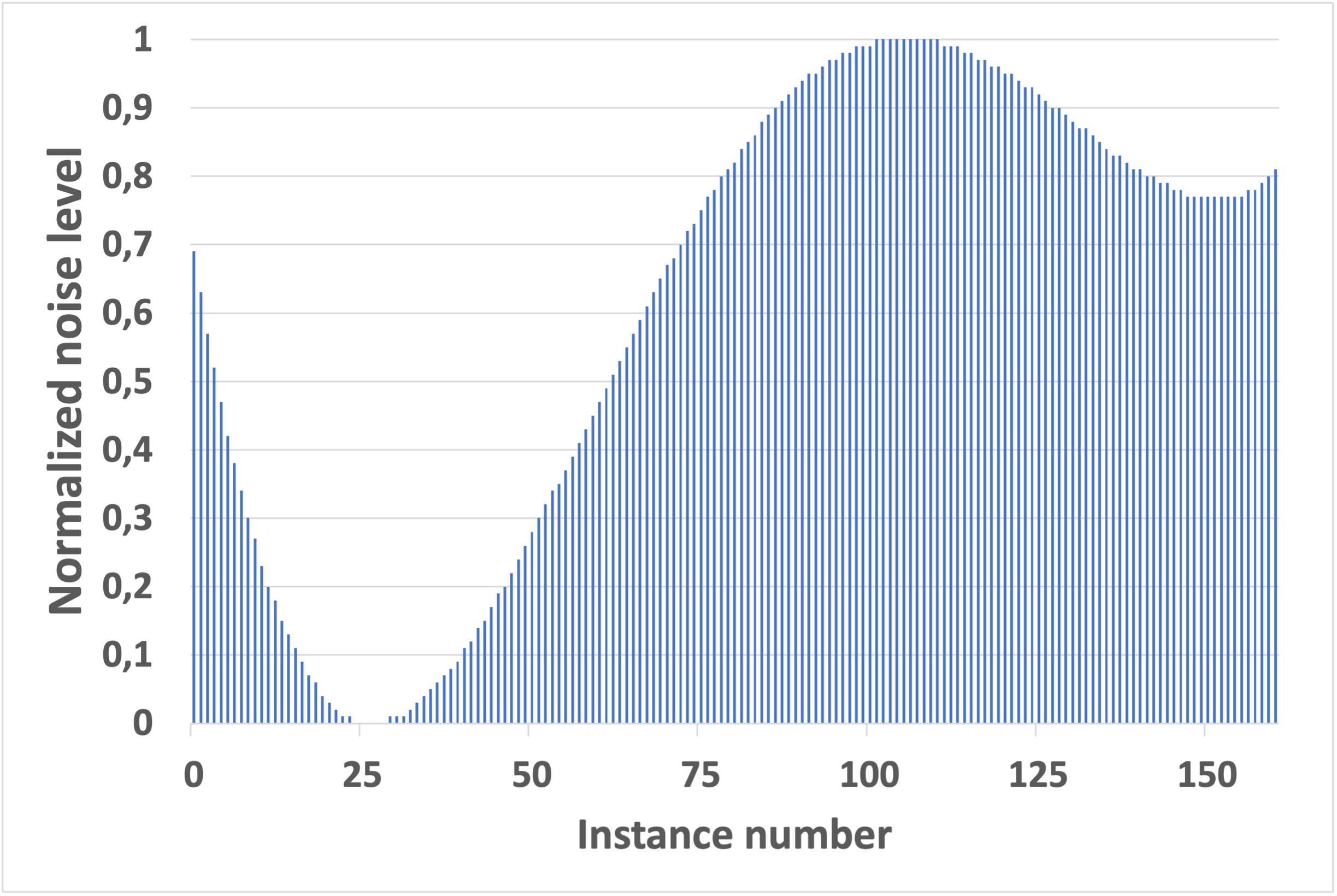}
    \caption{Noise distribution for the SKR trace: 160 noise level values, one for each sample in the trace. Each sample contains one random utterance (from the 12 classes) on which noise is added with the level specified according to the distribution of the ambient noise during a regular day~\cite{flamme2012typical}.}
    \label{fig:noise}
\end{figure}

Opportunities for dynamic approximation in SKR come with a naturally-varying level of background noise. For instance, it has been show that when different levels of noise are present, a different complexity of a DL model is needed to successfully recognize spoken keywords~\cite{machidon2022keyword}. In our experiments we examine how the adaptation strategies developed in Section~\ref{sec:strategies} cope with time-varying noise levels. For this, we first construct a trace consisting of 160 word utterances from a previously unseen part of the GSC dataset mixed with time-varying white noise whose level corresponds to the level measured in a realistic environment over 24 hours~\cite{flamme2012typical} (Figure~\ref{fig:noise}). The trace is then used for SKR with our mobile DL model, while a Mobiprox's adaptation strategy decides on which approximation configuration to use at subsequent inference step. Unlike in the HAR experiment, in this experiment there is no notion of ``state'' (e.g. a period of time during which a user is likely to keep performing the same activity), rather, keywords are randomly distributed in the trace. Thus, we do not evaluate the state-based adaptation method, but focus on the confidence-based adaptation.

\begin{table}[!htbp]
    \caption{Inference accuracy and energy consumption on the spoken keyword recognition task for a network introduced in~\cite{tang2017honk}.}
    \begin{center}
    \begin{tabular}{llll}
Adaptation  & Incr. & Accuracy & Relative Energy \\ \hline
Non-approximated & - & 0.96     & 1.0        \\ 
Confidence-based    & Expon.   & 0.96     & 0.852        \\ 
Confidence-based         & Linear   & 0.96     & 0.852        \\ 
\end{tabular}
    \end{center}
\label{tab:all_engines_sound}
\end{table}

We run the above trace on ASUS Tinkerboard S connected to a Monsoon power meter. We run both the original compact network from~\cite{tang2017honk} and the same network dynamically approximated with two flavors of our confidence-based adaptation scheme (with a linear and an exponential increase in approximation level). The results are shown in Table~\ref{tab:all_engines_sound}. Both the original network and the two flavors of the approximated network achieve the same accuracy $96.3\%$, while Mobiprox adaptation leads to 15\% system-wide energy savings. 


\blue{
\subsection{Smartphone-based adaptation in an unscripted scenario}
\label{sec:evaluation_unscripted}

We investigate how Mobiprox performs on a battery-powered commodity Android phone when the scenario of use is not prescribed. On all ten phones Mobiprox successfully ran real-time adaptation and the inference of \mnUci{} for human activity recognition. In addition, we re-ran the collected sensor traces on the same phone with both state-based and confidence-based approximation strategy employed. Analyzing the logs we have not observed any discrepancies (in terms of inference delay or inferred class mismatch) between on-device sampling and inference, and trace-based inference, confirming that Mobiprox affords smooth real-time approximation of mobile deep learning. 




\begin{table}[!htbp]
     \caption{Average relative accuracy (agreement with baseline) and relative energy consumption on 10 user traces collected in an unscripted scenario for MobileNet-V2 trained on the UCI-HAR dataset. Standard deviations across users are in parenthesis.}
     \begin{center}
     \begin{tabular}{llll}

 Adaptation  & Incr. & Agreement w. baseline & Rel. energy\\ \hline
 Confidence-based    & Expon.   & 0.83 (0.04)     & 0.85 (0.01)        \\ 
 State-based ($V_{L}=2$)         & Linear   & 0.91 (0.02)    & 0.88 (0.01)        \\ 
 \end{tabular}
     \end{center}
 \label{tab:unscripted}
 \end{table}

In Table~\ref{tab:unscripted} we compare the inference performance and energy savings for the confidence-based adaptation engine with exponential increase of approximation and the state-driven adaptation strategy with the linear increase. Since the experiments were unscripted and we do not have the ground truth labels, we show the relative accuracy (i.e., the agreement with the non-approximated baseline model) and relative energy consumption (i.e.,  compared to the consumption of the non-approximated model). From the table we observe that the state-based adaptation engine achieves a higher average agreement with the baseline non-approximated model -- 91\%, while consuming 12\% less energy than the baseline. The confidence-based engine allows for more energy savings -- up to 15\% -- but with the downside of a lower agreement \mbox{with the non-approximated network -- 83\%.}

To further understand the functioning of the approximation adaptation strategy, in Figure~\ref{fig:adaptation-timeline-bestenergy} we show an example of the adaptation timeline for one of the user traces collected in this experiment. The black dotted line presents the adaptation, where the higher approximation configuration number (right y-axis) indicates a higher level of approximation. We compare the activities inferred (left y-axis) by the baseline non-approximated model (green dots) with the activities inferred by the currently used approximation configuration. Should these differ, we plot a red triangle indicating the mismatched activity inferences. Finally, we show the cumulative energy savings (compared to the non-approximated model and normalised to the graph dimensions) extrapolated from the currently used configuration and the precise energy measurements from the ASUS Tinkerboard. 

The figure confirms that the Mobiprox adaptation strategy harnesses the \textit{accuracy-energy consumption} trade-off points determined during the tuning phase. The state-driven strategy remains cautious (i.e. uses more accurate configurations) when a user performs dynamic activities that are more difficult to classify (e.g. walking, walking up or down the stairs), thus when long periods of uniform classification results are not present. The strategy jumps to more aggressive approximation configurations when a user lingers in an easier-to-classify static activities (e.g. standing, lying, sitting). In terms of the mismatches with the baseline, we observe that the differences are often in individual timesteps (i.e., they agree soon again) and that diverging prediction are between similar activities (walk upstairs vs walking and sitting vs lying).

\begin{figure}[!h]
     \centering
     \includegraphics[width=\linewidth]{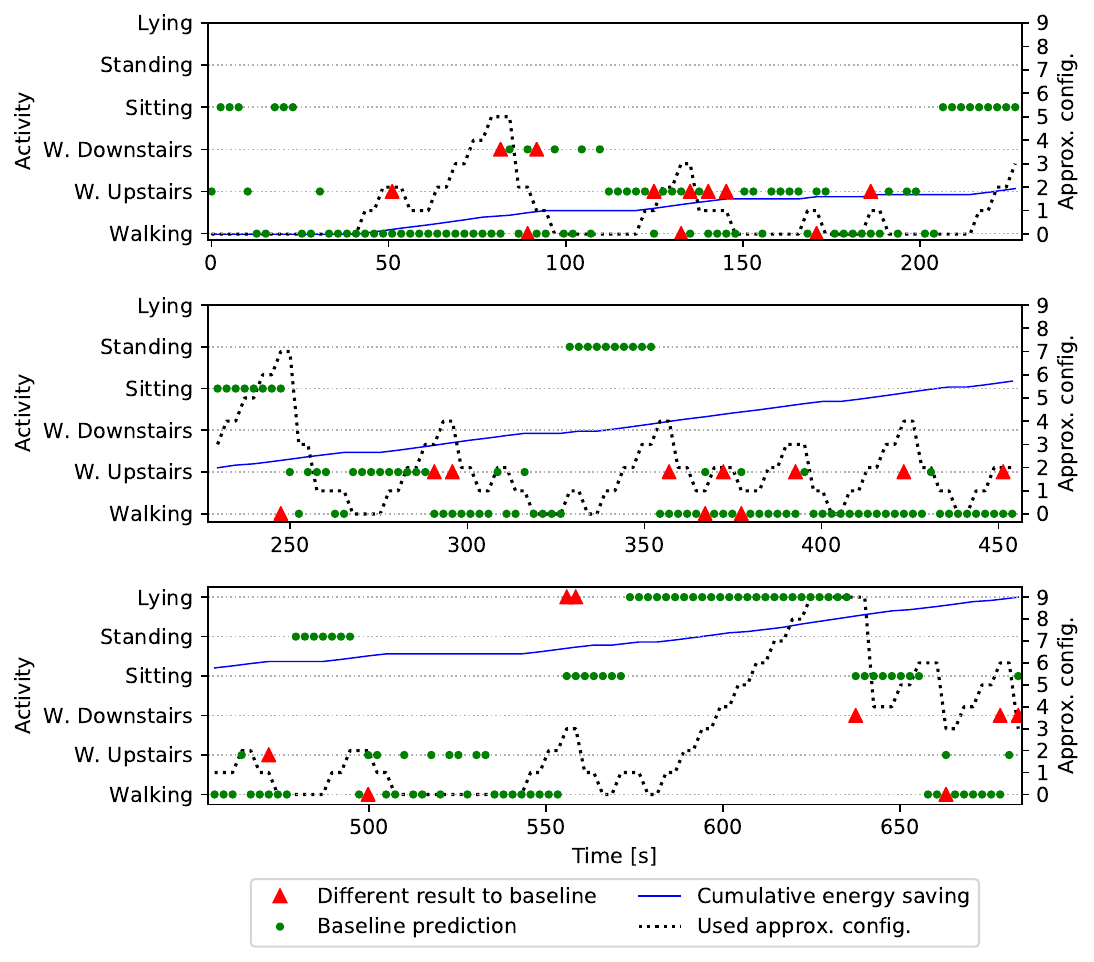}
     \caption{
     State-driven adaptation timeline with linear increase of approximation. }
     \label{fig:adaptation-timeline-bestenergy}
 \end{figure}

}

%% file: tex/discussion.tex
\section{Discussion and Limitations}
\label{sec:discussion}







Static training-time optimization is stifling further proliferation of mobile DL. Mobiprox builds upon the existing efforts towards dynamic DL optmisation~\cite{yu2020any, yu2018slimmable,laskaridis2020spinn,10.5555/3294771.3294979}, yet differs from them in three important ways: 
\textit{i)} being implemented at the compiler-level and exposed as an end-to-end pipeline, Mobiprox is not limited to a particular network architecture, supports various layer types present in NNs, \blue{supports models written in either PyTorch or Keras}, and does not require any previous involvement from a data scientist designing and training the network; \textit{ii)} Mobiprox supports quantization, perforated convolutions, and filter sampling, but is created to be easily extensible to a wide range of approximation techniques; \textit{iii)} Mobiprox allows context-dependent runtime adaptation of the approximation level; while we include two adaptation algorithms with the Mobiprox codebase, virtually any policy can be used. To facilitate future research and usage, we release Mobiprox as open-source software\footnote{\url{https://gitlab.fri.uni-lj.si/lrk/mobiprox/}}. 

Mobiprox, is subject to certain limitations. First, despite implementing an on-device profiler to better gauge the effect of different approximations on the QoS loss, Mobiprox cannot provide guarantees that the expected QoS will indeed be achieved, nor can it predict the maximal expected QoS loss on yet-to-be-seen data. Somewhat related is the issue of potentially reduced reliability of approximated models. While compression can, in certain situations, improve the generalizability of a model~\cite{giles1994pruning}, different compression levels can lead to widely varying reliability outcomes~\cite{cygert2021robustness}. 

Second, our measurements show the maximum speedup Mobiprox achieves on a mobile device remains relatively modest at $1.25\times$, while the same NN architecture achieves twice the speedup on a server. This discrepancy likely stems from the lack of optimised support for running (approximate) deep learning on mobile devices. The goal of the  prototype version of Mobiprox presented in this paper is to, for the first time, demonstrate dynamic mobile DL approximation adaptation. To unlock further benefits, we plan to examine integration with mobile DL compiler stacks that are already hand-optimized by large engineering teams in production environments, such as TVM~\cite{chen2018tvm}, Pytorch Mobile, or TF Lite\footnote{This is also the key reason why a direct comparison between Mobiprox and current mobile DL compression implementations (e.g. quantization in TFLite) is impossible -- neither do these approaches provide dynamic approximation adaptation, nor is Mobiprox optimized for performance.}. Since the approximations in Mobiprox reduce both the number of compute operations and memory loads and stores, the performance improvements of these approximations should seamlessly translate, if efficient library and compiler implementations listed above are used. Not only would this likely lead to improved speedup gains, but would also ameliorate the need for time-inefficient development of custom approximable tensor runtimes for various architectures. 



Third, Mobiprox is general and can be applied to any neural network architecture, yet, the richness of the approximate configurations and the efficiency of the approximation depends on the presence of convolutional layers in the network. Mobiprox specifically targets these layers with convolution perforation and filter sampling approximations, as convolutional layers tend to consume the majority of computational time and energy in mobile neural networks~\cite{li2018deeprebirth}. For non-convolutional layers, Mobiprox allows only one type of approximation -- half-precision quantization -- leading to at most $2^L$ possible approximation configurations in an L-layer network. To expand the range of approximation techniques, in future we plan to investigate the integration of dynamic pruning~\cite{10.5555/3294771.3294979} of fully-connected layers in Mobiprox. Techniques requiring up-front modification or specialized training to support approximation, such as Slimmable neural networks~\cite{yu2018slimmable}, remain unsuitable, as with Mobiprox we provide a service that allows the integration of pre-built networks oblivious to approximation, for instance, those acquired through Google Cloud AutoML, into mobile apps.

Finally, with respect to Mobiprox's adaptation algorithms (Section~\ref{sec:strategies}), these were developed for commonly encountered situations where the context does not fluctuate rapidly. Such behavior is present in numerous domains, including two examined in the previous section -- the human activity recognition where an activity a person is performing often stays the same over a certain time period, and the spoken keyword recognition, where the background noise gradually changes throughout the day. However, harnessing the slow-changing nature of many real-world phenomena, our adaptation strategies may not be suitable for tasks such as anomaly detection where sudden changes of the target phenomena are expected~\cite{pang2021deep}. Note that this does not restrict the general domain in which Mobiprox can be applied. Indeed, with minor modifications, the strategies presented in this paper can be used for adapting approximation of models built for tracking objects in live video~\cite{8003302}, for instance.

%% file: tex/conclusion.tex
\section{Conclusion}
\label{sec:conclusion}

In this paper we introduced Mobiprox, to the best of our knowledge, the first end-to-end framework that enables dynamically adaptable, rather than static, approximation of mobile DL. Furthermore, Mobiprox works with arbitrary architectures, even with networks not initially designed with approximation in mind. To accomplishing this, we first implemented low-level support for approximate computing on mobile CPU and GPUs through compiler-level primitives. We then integrated the heterogeneous compilation infrastructure, the approximate configuration search framework, and our novel profiler into an Android-ready end-to-end approximate configuration search, selection, and compilation pipeline. We ran different deep learning architectures 
through the pipeline and 
demonstrated that Mobiprox identifies approximation configurations that enable a trade-off between inference accuracy and energy consumption. Finally, we implemented approximation adaptation strategies for dynamic selection of energy-preserving DL configurations while ensuring that the quality of the resulting classification is not hurt. Experiments in human activity and spoken keyword recognition domains demonstrate that the adaptation strategies successfully accommodate varying context, reducing the system-wide energy usage by 15\% in both domains, while sacrificing only 2\% of the accuracy in the HAR domain, and leading to no loss of accuracy in the spoken keyword recognition domain. 
